\documentclass{article}

\usepackage[preprint]{neurips_2023}

\usepackage[utf8]{inputenc} 
\usepackage[T1]{fontenc}    
\usepackage{hyperref}       
\usepackage{url}            
\usepackage{booktabs}       
\usepackage{amsfonts}       
\usepackage{nicefrac}       
\usepackage{microtype}      
\usepackage{xcolor}         

\usepackage{algorithm}
\usepackage{amsmath}
\usepackage{algpseudocode}
\usepackage{graphicx}
\usepackage{wrapfig}
\usepackage{subfigure}

\usepackage{amssymb}
\usepackage{mathtools}
\usepackage{amsthm}
\usepackage{bm}
\usepackage{amsmath}
\usepackage{diagbox}

\title{Uncertainty Quantification using Generative Approach}

\author{%
  Yunsheng Zhang \\
  \texttt{yunshzhang@gmail.com} \\
}

\newtheorem{theorem}{Theorem}

\begin{document}

\maketitle

\begin{abstract}
We present the Incremental Generative Monte Carlo (IGMC) method, designed to measure uncertainty in deep neural networks using deep generative approaches. IGMC iteratively trains generative models, adding their output to the dataset, to compute the posterior distribution of the expectation of a random variable. We provide a theoretical guarantee of the convergence rate of IGMC relative to the sample size and sampling depth. Due to its compatibility with deep generative approaches, IGMC is adaptable to both neural network classification and regression tasks. We empirically study the behavior of IGMC on the MNIST digit classification task.
\end{abstract}

\section{Introduction}
\label{sec:1}

Deep learning has found applications across many domains in recent years and has consistently achieved remarkable results \citep{deep_learning}. While deep neural networks excel in predictive accuracy across various tasks, their predictions are not devoid of errors.
\emph{Uncertainty} can be defined as a measure of the divergence between predicted values and their ground-truth values. 
Quantifying the uncertainty in deep neural network prediction is crucial in several research fields. For example, in tasks requiring high-risk decision-making or stringent safety standards, such as medical image analysis \citep{medical1, medical2, medical3} and autonomous vehicle control \citep{vehicle1, vehicle2}, uncertainty quantification can assist AI systems in assessing challenging situations. In reinforcement learning, uncertainty quantification can aid agents in improving their exploration strategies \citep{bayesian_dqn, count-based}. Uncertainty quantification is also widely used in active learning \citep{active1, active2}, explainable AI \citep{explain1}, and few-shot learning \citep{fewshot1, fewshot2}.

There are many possible sources of uncertainty in neural network prediction, including data uncertainty, model uncertainty \citep{uncertainty_source}, or out-of-distribution (OOD) of test data \citep{ood1, ood2}. Traditional deep learning methods cannot capture uncertainty. Various methods have been proposed to quantify uncertainty, such as A single deterministic network to predict uncertainty \citep{single1, single2, single3}. Augment the input data at test-time \citep{ttaug1, ttaug2}. The Bayesian method \citep{mc_dropout, bayes}. The ensemble method trains several models and combines their prediction to estimate uncertainty during inference \citep{ensemble1, ensemble2}.

This paper focuses on the uncertainty of models that predict a random variable's expectation or conditional expectation.
Note that classification (which provides probabilities for each class label as continuous values) and regression can be described as forms of conditional expectation learning. 
Given a set of independent and identically distributed (i.i.d.) samples $\mathcal{S}$ from a random variable $X$, our method, named \emph{Incremental Generative Monte Carlo} (IGMC), determines the posterior cumulative distribution function (CDF) for $\mu=\mathrm{E}[X]$, thereby quantifying the uncertainty.
Unlike Bayesian methods, our method's definition of the posterior CDF is not based on likelihoods or parameter priors but rather on a generative approach. A \emph{generative approach} is a rule or algorithm that converts any sample set into a generative model. The models and learning algorithms of variational autoencoder \citep{vae} and generative adversarial network \citep{gan} can be considered examples of generative approaches.

Compared to other methods, our method requires fewer modifications to the model and its learning algorithm. Notably, a classification model can naturally function as a generative model, obviating the need for changes when addressing classification tasks, which implies that the classification algorithm has the inherent ability to gauge classification uncertainty.
We theoretically and empirically show the distance between the posterior CDF produced by IGMC and the ground-truth posterior CDF.
Lastly, we empirically study IGMC's behavior on the MNIST digit classification task \citep{mnist}, using a convolutional neural network as the classification model.

\section{Preliminary}
\label{sec:2}

Let $X$ denote a random variable with the support $[0, 1]$, with an unknown expectation value $\mu=\mathbf{\mathrm{E}}[X]$. We draw $M$ independent samples from $X$, symbolized as $x_1, x_2, \ldots, x_M$, and the empirical estimate for $\mu$ is given by $\hat\mu = \frac{x_1 + x_2 + \ldots + x_M}{M}$.

One way to define uncertainty is by assessing the probability that $\mu$ and $\hat \mu$ deviate by a certain amount $t$. We can determine the upper limit of this uncertainty through Hoeffding's inequality:

\begin{equation}
\mathrm{Pr}(|\hat\mu-\mu|\geq t)\leq 2\exp(-2Mt^2)
\end{equation}

This inequality illustrates that the uncertainty shrinks exponentially as the sample size increases. Another commonly used Bayesian method interprets $\mu$ as a posterior distribution conditioned on $x_1, x_2, \ldots, x_M$. With a known form of parameter $\theta$, and its prior $\mathrm{Pr}(\theta)$ and likelihoods $\mathrm{Pr}(x|\theta)$, the Bayesian posterior probability is given as follow:

\begin{equation}
\mathrm{Pr}(\theta|x_1,\ldots,x_M)=\frac{\mathrm{Pr}(x_1, \ldots, x_M|\theta)}{\int\mathrm{Pr}(x_1, \ldots, x_M|\theta')\cdot \mathrm{Pr}(\theta')\mathrm{d}\theta'}\cdot \mathrm{Pr}(\theta)
\end{equation}

Thus, the posterior probability of $\mu$ can be obtained by:

\begin{equation}
\mathrm{Pr}(\mu|x_1,\ldots,x_M)=\int\mathrm{Pr}(\mu|\theta)\mathrm{Pr}(\theta|x_1,\ldots,x_M)\mathrm{d}\theta
\end{equation}

The uncertainty can be measured by the dispersion of posterior distribution, such as the standard deviation of $\mathrm{Pr}(\mu|x_1,\ldots,x_M)$.

\section{Generative approach as uncertainty quantifier}
\label{sec:3}

In this section, we present a method for quantifying uncertainty using generative approaches. We first give the definition of the generative approach. The generative approach, denoted as $\Phi$, can transform any samples drawn from $X$ into a generative model. The generative model represents a random variable with a support range identical to $X$. Note that we will focus only on one-dimensional $X$ in this section.

Consider a Bernoulli trial as an illustration. For any $M$ observations represented by $x_1,\ldots,x_M$, let $a$ denote the number of success events among them. Then $\Phi(x_1,\ldots,x_M):=\mathrm{Bern}(a/M)$ is a well-defined generative approach. For notational simplicity, we assume $\Phi$ is deterministic, and our results can generalize to the case that $\Phi$ is nondeterministic.

\subsection{Definition of Incremental Generative Distribution Function}
\label{sec:3.1}

The intuition behind using a generative approach to compute the posterior distribution of $\mathrm{E}[X]$ is as follows: as the sample size of $X$ grows, the uncertainty of $\mathrm{E}[X]$ will decrease. Suppose we have a good generative approach $\Phi$ that can effectively describe the relationship between observations and distributions. In that case, we can generate enough samples by utilizing the generative approach $\Phi$ to represent the posterior distribution in a simple and computable form.

Using mathematical notation, considering calculation of the posterior CDF $\mathrm{Pr}(\mu<t|\mathcal{S})$ of $\mu$, we use $\mathcal{S}$ to represent $x_1,\ldots,x_M$ for short. We can expand the function by applying the total probability formula:

\begin{equation}
\mathrm{Pr}(\mu<t|\mathcal{S})=\int_0^1\mathrm{d}y_1\mathrm{Pr}(y_1|\mathcal{S})\mathrm{Pr}(\mu<t|\mathcal{S},y_1)    
\end{equation}

Here $\mathrm{Pr}(y_1|\mathcal{S})$ is the probability density function of $\Phi(\mathcal{S})$. We can then proceed to apply the full probability formula to $\mathrm{Pr}(\mu<t|\mathcal{S},y_1)$:

\begin{equation}
\mathrm{Pr}(\mu<t|\mathcal{S},y_1)=\int_0^1\mathrm{d}y_2\mathrm{Pr}(y_2|\mathcal{S},y_1)\mathrm{Pr}(\mu<t|\mathcal{S} ,y_1,y_2)    
\end{equation}

Here $\mathrm{Pr}(y_2|\mathcal{S},y_1)$ is the probability density function of $\Phi(\mathcal{S},y_1)$. Repeating this process for $H$ times, we define:

\begin{equation}
F^{\Phi}_H(t|\mathcal{S})=\underbrace{\int_0^1\mathrm{d}y_1\mathrm{Pr}(y_1|\mathcal{S})\cdots\cdots\int_0^1\mathrm{d}y_h \mathrm{Pr}(y_H|\mathcal{S},y_1,\ldots,y_{H-1})}_{\mathrm{applying\ law\ of\ total\ probability\ for}\ H\ \mathrm{times}}\mathbb{I} (\frac{\sum x+\sum y}{M+H}<t)
\label{eq:6}
\end{equation}

We substituted the posterior probability with an indicator function in the last item. Since $\mathrm{Pr}(\mu<t|\mathcal{S},y_1,...,y_H)$ approaches $\mathbb{I}(\frac{\sum x+\sum y}{M+H}<t)$ arbitrarily close as $H\to\infty$, we define

\begin{equation}
F^{\Phi}_{\infty}(t|\mathcal{S})=\lim_{H\to\infty}F^{\Phi}_H(t|\mathcal{S})
\label{eq:7}
\end{equation}

Then $F^{\Phi}_\infty(t|\mathcal{S})$ should correspond to the posterior CDF of $\mathrm{E}[X]$ that we aim to obtain. In the above formulation, only the generative approach $\Phi$ needs to be provided. Therefore, Eq. \eqref{eq:6} and \eqref{eq:7} can serve as the definition for the posterior CDF based on the specified generative approach $\Phi$. We term \(F^{\Phi}_\infty(t|\mathcal{S})\) as the \emph{Incremental Generative Distribution Function} (IGDF).

From the Bayesian perspective, the definition of posterior probability requires the prior and the likelihood. Interestingly, our method does not require parameters prior explicitly. Notably, this absence is not contradictory because $\Phi$ already has the information about the parameter prior and likelihood inherently. This claim is supported by the fact that $F^{\Phi}_{\infty}(t|\mathcal{S})$ matches the posterior distribution from the Bayesian method when $\Phi(\mathcal{S})$ consistently follows the Bayesian posterior predictive distribution for any sample set $\mathcal{S}$.

\subsection{Incremental Generative Monte Carlo}

\begin{algorithm}[t]
    \centering
    \caption{Incremental Generative Monte Carlo}\label{alg:IGMC}
    \begin{algorithmic}[1]
        \State Given generative approach $\Phi$; i.i.d. samples $\{x_1,\ldots,x_M\}$ from $X$; sample size $N$; sampling depth $H$.
        \For{$n = 1,\ldots,N$}
            \State Initialize the dataset $\mathcal{S}_n\leftarrow\{x_1,\ldots,x_M\}$
            \For{$h = 1,\ldots,H$}
                \State get generative model $\gamma\leftarrow\Phi(\mathcal{S}_n)$
                \State sample $y_h\sim\gamma$
                \State add $y_h$ into $\mathcal{S}_n$
            \EndFor
            \State set $\mu_n\leftarrow\frac{\sum x_i+\sum y_h}{M+H}$
        \EndFor
        \State set $\hat{F}_{H,N}^\Phi(t|\mathcal{S})\leftarrow\sum_{n=1}^N \mathbb{I}(\mu_n \leq t)$
        \State \textbf{Return} $\hat{F}_{H,N}^\Phi(\cdot|\mathcal{S})$
    \end{algorithmic}
\end{algorithm}

In Section \ref{sec:3.1}, we define the IGDF $F_\infty^\Phi(t|\mathcal{S})$ based on the generative approach $\Phi$. Intuitively, for a large $H$, $F_H^\Phi(t|\mathcal{S})$ is an approximation of $F_\infty^\Phi(t|\mathcal{S})$.
However, obtaining the closed form of either $F_\infty^\Phi(t|\mathcal{S})$ or $F_H^\Phi(t|\mathcal{S})$ may not always be possible for a given $\Phi$.
Since each $y_h$ in Eq.~\eqref{eq:6} can be sampled from $\Phi(\mathcal{S},y_1 ,\ldots,y_{h-1})$, we can approximate $F_H^\Phi(t|\mathcal{S})$ using the Monte Carlo method. We present the complete algorithm, which we call \textit{Incremental Generative Monte Carlo} (IGMC), in Algorithm~\ref{alg:IGMC}. The $\hat F_{H,N}^\Phi(\cdot| \mathcal{S})$ returned by IGMC is the Monte Carlo estimation of $F_H^\Phi(\cdot|\mathcal{S})$ with a sample size of $N$.

\begin{figure*}[t]
\centering

\subfigure[Bernoulli cases. The samples are set to have $M=9$ and $a=4$.]{
\label{fig:bern}
\hspace{-25pt}
\includegraphics[width=\linewidth/3+10pt]{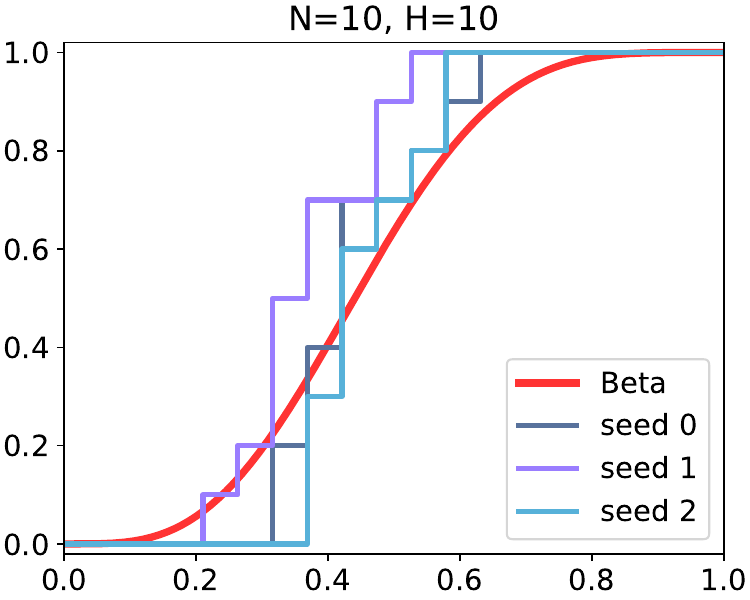}
\includegraphics[width=\linewidth/3+10pt]{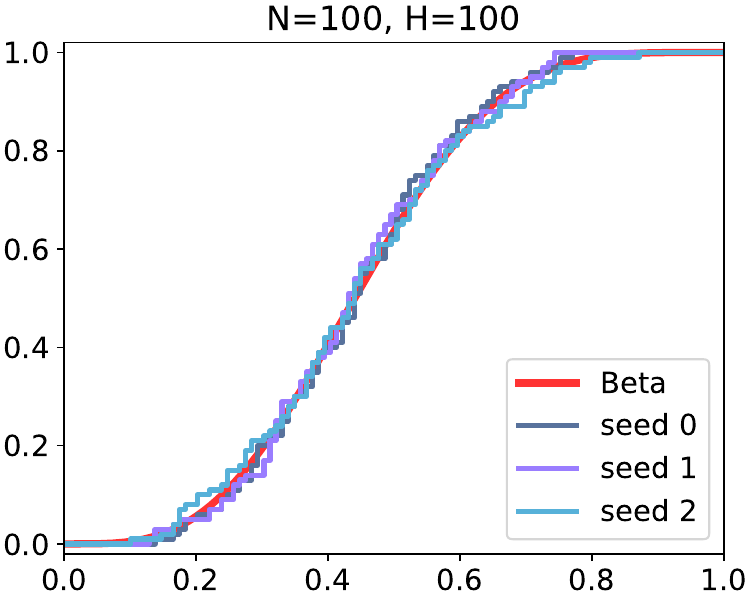}
\includegraphics[width=\linewidth/3+10pt]{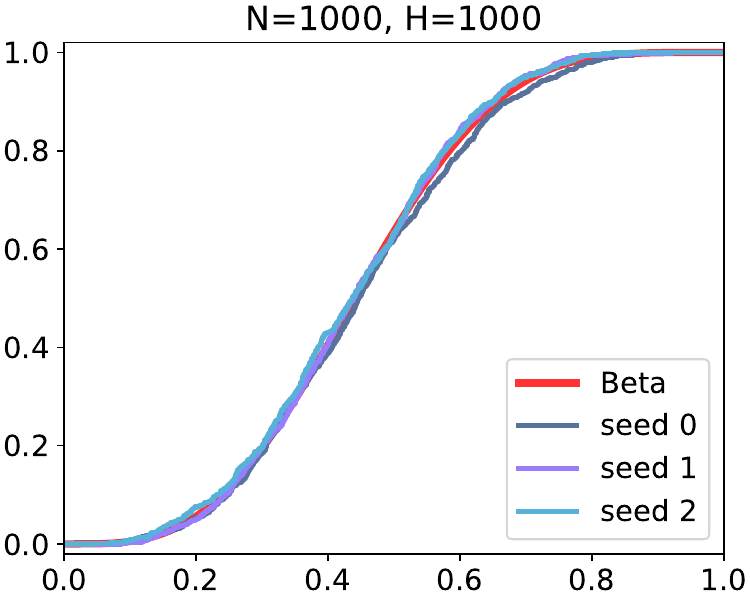}
}
\quad
\subfigure[Exponential cases. The samples are set to have $M=50$ and $\hat\mu=2.0$.]{
\label{fig:exp}
\hspace{-25pt}
\includegraphics[width=\linewidth/3+10pt]{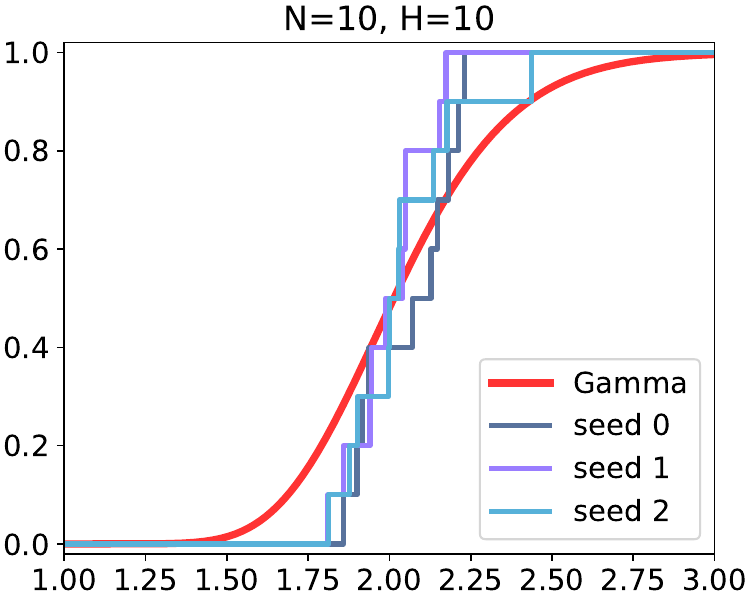}
\includegraphics[width=\linewidth/3+10pt]{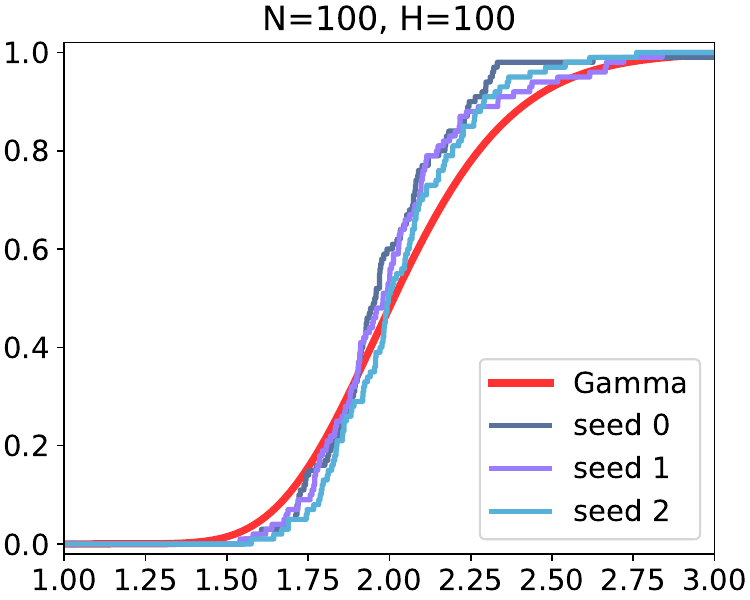}
\includegraphics[width=\linewidth/3+10pt]{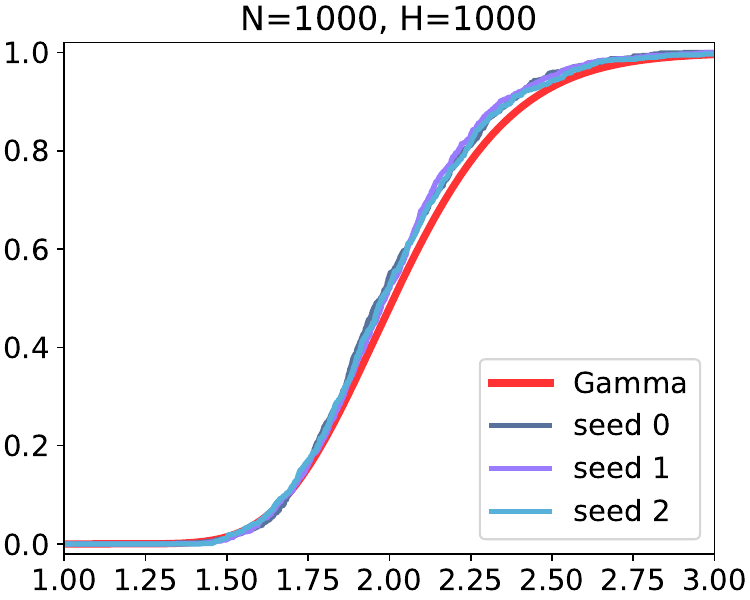}
}

\caption{Experimental curves of IGMC for specific $N$ and $H$ values.}
\label{fig:mnist}
\end{figure*}

We empirically test IGMC for situations where $X$ follows Bernoulli and exponential distribution. For the cases where $X$ follows the Bernoulli distribution, we set $\Phi(x_1,\ldots,x_M)\coloneqq\mathrm{Bern}(a/M)$, where $a$ is the number of success events. For the cases that $X$ follows the exponential distribution, we set $\Phi(x_1,\ldots,x_M)\coloneqq\mathrm{Exponential}(1/\hat\mu)$, with $\hat\mu$ being the average of $x_1,\ldots,x_M$. All experiments run with three different random seeds. The configurations are $N=10, H=10$; and $N=100, H=100$; and $N=1000, H=1000$. Additionally, we plotted the posterior distribution of $\mu$ by the Bayesian method for reference, that is, $\mathrm{Beta}(a, M-a)$ and $\mathrm{Gamma}(M, M\hat\mu)$. The resulting curves of $\hat F_{H,N}^\Phi$ are shown in Fig.~\ref{fig:bern} and Fig.~\ref{fig:exp}.

The figure indicates that when $N$ and $H$ are greater than a hundred, the $\hat F_{H,N}^\Phi$ produced by IGMC aligns closely with the distribution derived from the Bayesian method. When $X$ follows the Bernoulli distribution, the Bayesian posterior predictive distribution follows $ \mathrm{Bern}(a, M-a)$. Thus, as $N$ and $H$ increase, the $\hat F_{H,N}^\Phi$ will converge to $\mathrm{Beta}(a, M-a)$. However, when $X$ follows an exponential distribution, the $\hat F_{H,N}^\Phi$ does not match the Gamma distribution, even as $N$ and $H$ approach infinity. This discrepancy is because the Bayesian posterior predicts a Lomax distribution instead of $\mathrm{Exp}(1/\hat{\mu})$ in this situation.

In addition to empirical evaluations, it is crucial to verify the convergence of $\hat F^\Phi_{H,N}$ to the IGDF $F^\Phi_\infty$ and determine its rate of convergence. The following theorem provides a theoretical guarantee under some conditions.

\begin{theorem}
\label{theorem:1}
If support of $X$ is $[0,1]$, and $\Phi$ satisfies $\mathrm{E}\left[\Phi(x_1,\ldots,x_M)\right]=\sum_{i=1}^M x_i/M$ for $\forall M \in \mathbb{N}^+$ and $\forall x_1,\ldots,x_M\in [0,1]^M$. Then the expectation of $L^1$ distance between $F^\Phi_\infty$ and $\hat F^\Phi_{H,N}$ is $O(\sqrt{\frac{1}{N}}+\sqrt{\frac{1}{H}})$, where $\hat F^\Phi_{H,N}$ is the posterior CDF returned by IGMC (Algorithm~\ref{alg:IGMC}) with sampling depth $H$ and sample size $N$.
\end{theorem}

\textbf{Proof:} See Appendix~\ref{appendix:proof}.

\section{Deep Incremental Generative Monte Carlo}
\label{sec:4}

\begin{algorithm}[t]
    \centering
    \caption{Deep Incremental Generative Monte Carlo for Classification Tasks}\label{alg:DEEP}
    \begin{algorithmic}[1]
        \State Given deep classification learning algorithm $\Phi$; Training dataset $\mathcal{D}$; sample size $N$; sampling depth $H$; Number of categories $K$; Input $x$ for classification and uncertainty quantification.
        \For{$n = 1,\ldots,N$}
            \State Initialize the dataset $\mathcal{D}_n\leftarrow\mathcal{D}$
            \For{$h = 1,\ldots,H$}
                \State initialize neural network parameters $\theta$
                \State train $\theta$ on dataset $\mathcal{D}_n$ by $\Phi$ until converge.
                \State compute $\bm{p}\leftarrow f_\theta(x)$ \label{line7}
                \State sample a label $y_h\sim \bm{p}$
                \State add input-label pair $(x,y_h)$ into $\mathcal{D}_n$
            \EndFor
            \For{$k = 1,\ldots,K$}
            \State set $\mu_{n,k}\leftarrow\sum_{h=1}^H\mathbb{I}(y_h=k)/H$
            \EndFor
        \EndFor 
        \For{$k = 1,\ldots,K$}
        \State set $\hat{F}_{H,N}^\Phi(t,k|\mathcal{D})\leftarrow\sum_{n=1}^N \mathbb{I}(\mu_{n,k}\leq t)$
        \EndFor
        \State \textbf{Return} $\hat{F}_{H,N}^\Phi(\cdot,\cdot|\mathcal{D})$
    \end{algorithmic}
\end{algorithm}

In this section, we extend Algorithm~\ref{alg:IGMC} to quantify the uncertainty in the predictions of deep neural networks. This is achieved by substituting $\Phi$ with deep generative methods such as VAE~\citep{vae}, GAN~\citep{gan}, and Diffusion~\citep{ddpm}. For a discrete target space, a feed-forward neural network with a softmax activation function in the final layer can serve as $\Phi$. Different from Section~\ref{sec:3}, the deep neural network generally requires an input $x$. Thus, we must add $(x,y_h)$ pair to the dataset $\mathcal{D}_n$ rather than only $y_h$.

We take the classification task with $K$  categories as an example. The pseudo-code is shown in Algorithm~\ref{alg:DEEP}. Note in line \ref{line7}, $f_\theta(x)$ denotes the neural network's output given input $x$ and parameter $\theta$. $\bm{p}$ is a $K$-dimensional vector where $\bm {p}_k$ represents the probability that $x$ belongs to class $k$. Let $P_k$ be a random variable signifying the probability of input $x$ belonging to class $k$. The outcome $\hat{F}_{H,N}^\Phi(t,k|\mathcal{D})$ of Alg~\ref{alg:DEEP} represents the estimated posterior probability of $P_k\leq t$.

\begin{table}[b]
\hspace{-1.5cm}
\begin{tabular}{l|llllllllll}
\hline
{\small Labels} &
\begin{minipage}[b]{0.04\linewidth}{\vspace{.05cm}\includegraphics{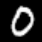}}\end{minipage} &
\begin{minipage}[b]{0.04\linewidth}{\includegraphics{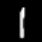}}\end{minipage} &
\begin{minipage}[b]{0.04\linewidth}{\includegraphics{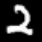}}\end{minipage} &
\begin{minipage}[b]{0.04\linewidth}{\includegraphics{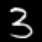}}\end{minipage} &
\begin{minipage}[b]{0.04\linewidth}{\includegraphics{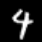}}\end{minipage} &
\begin{minipage}[b]{0.04\linewidth}{\includegraphics{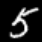}}\end{minipage} &
\begin{minipage}[b]{0.04\linewidth}{\includegraphics{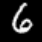}}\end{minipage} &
\begin{minipage}[b]{0.04\linewidth}{\includegraphics{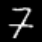}}\end{minipage} &
\begin{minipage}[b]{0.04\linewidth}{\includegraphics{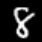}}\end{minipage} &
\begin{minipage}[b]{0.04\linewidth}{\includegraphics{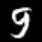}}\end{minipage} \\
\hline
0 & {\tiny $100.0 [\textcolor{red}{0.00}$]} & {\tiny $0.0 [\textcolor{red}{0.00}$]} & {\tiny $0.0 [\textcolor{red}{0.00}$]} & {\tiny $0.0 [\textcolor{red}{0.00}$]} & {\tiny $0.0 [\textcolor{red}{0.00}$]} & {\tiny $0.0 [\textcolor{red}{0.00}$]} & {\tiny $0.0 [\textcolor{red}{0.00}$]} & {\tiny $0.0 [\textcolor{red}{0.00}$]} & {\tiny $0.0 [\textcolor{red}{0.00}$]} & {\tiny $0.0 [\textcolor{red}{0.00}$]} \\
1 & {\tiny $0.0 [\textcolor{red}{0.00}$]} & {\tiny $100.0 [\textcolor{red}{0.00}$]} & {\tiny $0.0 [\textcolor{red}{0.00}$]} & {\tiny $0.0 [\textcolor{red}{0.00}$]} & {\tiny $0.0 [\textcolor{red}{0.00}$]} & {\tiny $0.0 [\textcolor{red}{0.00}$]} & {\tiny $0.0 [\textcolor{red}{0.00}$]} & {\tiny $0.1 [\textcolor{red}{0.00}$]} & {\tiny $0.0 [\textcolor{red}{0.00}$]} & {\tiny $0.0 [\textcolor{red}{0.00}$]} \\
2 & {\tiny $0.0 [\textcolor{red}{0.00}$]} & {\tiny $0.0 [\textcolor{red}{0.00}$]} & {\tiny $100.0 [\textcolor{red}{0.00}$]} & {\tiny $0.0 [\textcolor{red}{0.00}$]} & {\tiny $0.0 [\textcolor{red}{0.00}$]} & {\tiny $0.0 [\textcolor{red}{0.00}$]} & {\tiny $0.0 [\textcolor{red}{0.00}$]} & {\tiny $30.1 [\textcolor{red}{0.46}$]} & {\tiny $0.0 [\textcolor{red}{0.00}$]} & {\tiny $0.0 [\textcolor{red}{0.00}$]} \\
3 & {\tiny $0.0 [\textcolor{red}{0.00}$]} & {\tiny $0.0 [\textcolor{red}{0.00}$]} & {\tiny $0.0 [\textcolor{red}{0.00}$]} & {\tiny $100.0 [\textcolor{red}{0.00}$]} & {\tiny $0.0 [\textcolor{red}{0.00}$]} & {\tiny $0.0 [\textcolor{red}{0.00}$]} & {\tiny $0.0 [\textcolor{red}{0.00}$]} & {\tiny $6.7 [\textcolor{red}{0.13}$]} & {\tiny $0.0 [\textcolor{red}{0.00}$]} & {\tiny $8.8 [\textcolor{red}{0.13}$]} \\
4 & {\tiny $0.0 [\textcolor{red}{0.00}$]} & {\tiny $0.0 [\textcolor{red}{0.00}$]} & {\tiny $0.0 [\textcolor{red}{0.00}$]} & {\tiny $0.0 [\textcolor{red}{0.00}$]} & {\tiny $84.2 [\textcolor{red}{0.28}$]} & {\tiny $0.0 [\textcolor{red}{0.00}$]} & {\tiny $0.0 [\textcolor{red}{0.00}$]} & {\tiny $0.1 [\textcolor{red}{0.00}$]} & {\tiny $0.0 [\textcolor{red}{0.00}$]} & {\tiny $0.0 [\textcolor{red}{0.00}$]} \\
5 & {\tiny $0.0 [\textcolor{red}{0.00}$]} & {\tiny $0.0 [\textcolor{red}{0.00}$]} & {\tiny $0.0 [\textcolor{red}{0.00}$]} & {\tiny $0.0 [\textcolor{red}{0.00}$]} & {\tiny $0.0 [\textcolor{red}{0.00}$]} & {\tiny $100.0 [\textcolor{red}{0.00}$]} & {\tiny $0.0 [\textcolor{red}{0.00}$]} & {\tiny $0.0 [\textcolor{red}{0.00}$]} & {\tiny $0.0 [\textcolor{red}{0.00}$]} & {\tiny $3.5 [\textcolor{red}{0.08}$]} \\
6 & {\tiny $0.0 [\textcolor{red}{0.00}$]} & {\tiny $0.0 [\textcolor{red}{0.00}$]} & {\tiny $0.0 [\textcolor{red}{0.00}$]} & {\tiny $0.0 [\textcolor{red}{0.00}$]} & {\tiny $0.0 [\textcolor{red}{0.00}$]} & {\tiny $0.0 [\textcolor{red}{0.00}$]} & {\tiny $100.0 [\textcolor{red}{0.00}$]} & {\tiny $0.0 [\textcolor{red}{0.00}$]} & {\tiny $0.0 [\textcolor{red}{0.00}$]} & {\tiny $0.0 [\textcolor{red}{0.00}$]} \\
7 & {\tiny $0.0 [\textcolor{red}{0.00}$]} & {\tiny $0.0 [\textcolor{red}{0.00}$]} & {\tiny $0.0 [\textcolor{red}{0.00}$]} & {\tiny $0.0 [\textcolor{red}{0.00}$]} & {\tiny $2.4 [\textcolor{red}{0.03}$]} & {\tiny $0.0 [\textcolor{red}{0.00}$]} & {\tiny $0.0 [\textcolor{red}{0.00}$]} & {\tiny $62.0 [\textcolor{red}{0.54}$]} & {\tiny $0.0 [\textcolor{red}{0.00}$]} & {\tiny $0.0 [\textcolor{red}{0.00}$]} \\
8 & {\tiny $0.0 [\textcolor{red}{0.00}$]} & {\tiny $0.0 [\textcolor{red}{0.00}$]} & {\tiny $0.0 [\textcolor{red}{0.00}$]} & {\tiny $0.0 [\textcolor{red}{0.00}$]} & {\tiny $0.0 [\textcolor{red}{0.00}$]} & {\tiny $0.0 [\textcolor{red}{0.00}$]} & {\tiny $0.0 [\textcolor{red}{0.00}$]} & {\tiny $0.7 [\textcolor{red}{0.00}$]} & {\tiny $100.0 [\textcolor{red}{0.00}$]} & {\tiny $2.0 [\textcolor{red}{0.04}$]} \\
9 & {\tiny $0.0 [\textcolor{red}{0.00}$]} & {\tiny $0.0 [\textcolor{red}{0.00}$]} & {\tiny $0.0 [\textcolor{red}{0.00}$]} & {\tiny $0.0 [\textcolor{red}{0.00}$]} & {\tiny $13.2 [\textcolor{red}{0.23}$]} & {\tiny $0.0 [\textcolor{red}{0.00}$]} & {\tiny $0.0 [\textcolor{red}{0.00}$]} & {\tiny $0.4 [\textcolor{red}{0.00}$]} & {\tiny $0.0 [\textcolor{red}{0.00}$]} & {\tiny $85.7 [\textcolor{red}{0.24}$]} \\
\hline
\end{tabular}

\caption{Results for images from MNIST test dataset.}
\label{tab:1}
\end{table}

We evaluate Algorithm~\ref{alg:DEEP} on the MNIST digit classification task \citep{mnist}. The experimental model is a two-layer convolution neural network with 16 channels and ReLU activation functions, ending with a linear softmax classification layer. The initial dataset is the MNIST training dataset. The models are trained by the SGD optimizer with cosine learning rate decay. The learning rate is initiated to $0.002$, and the momentum is set to $0.9$. Convergence is deemed after $30$ training epochs.

We randomly selected some images from the MNIST test dataset to evaluate IGMC. The results are presented in Table~\ref{tab:1}. The black numbers in the table represent the percentage probability of a prediction belonging to the corresponding category. The number in red inside the square brackets is $4\sigma^2$, where $\sigma^2$ is the variance of the $\hat F_{H,N}$ for the corresponding category returned by IGMC. This red number can be viewed as a measure of uncertainty, ranging from $0.0$ to $1.0$. A larger value indicates greater uncertainty.

We also test different rotation angles of an image and some letter images from the EMNIST dataset \citep{emnist} to simulate the situation where the test data are out of distribution. The results are shown in Table~\ref{tab:2} and Table~\ref{tab:3}.

\begin{table}[h]
\begin{center}
\begin{tabular}{l|llllllll}
\hline
{\small Labels} &
\begin{minipage}[b]{0.04\linewidth}{\vspace{.05cm}\includegraphics{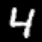}}\end{minipage} &
\begin{minipage}[b]{0.04\linewidth}{\includegraphics{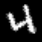}}\end{minipage} &
\begin{minipage}[b]{0.04\linewidth}{\includegraphics{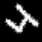}}\end{minipage} &
\begin{minipage}[b]{0.04\linewidth}{\includegraphics{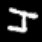}}\end{minipage} &
\begin{minipage}[b]{0.04\linewidth}{\includegraphics{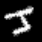}}\end{minipage} &
\begin{minipage}[b]{0.04\linewidth}{\includegraphics{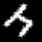}}\end{minipage} &
\begin{minipage}[b]{0.04\linewidth}{\includegraphics{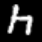}}\end{minipage} &
\begin{minipage}[b]{0.04\linewidth}{\includegraphics{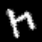}}\end{minipage} \\
\hline
0 & {\tiny $0.0 [\textcolor{red}{0.00}$]} & {\tiny $0.0 [\textcolor{red}{0.00}$]} & {\tiny $0.6 [\textcolor{red}{0.01}$]} & {\tiny $0.0 [\textcolor{red}{0.00}$]} & {\tiny $0.0 [\textcolor{red}{0.00}$]} & {\tiny $0.0 [\textcolor{red}{0.00}$]} & {\tiny $6.1 [\textcolor{red}{0.19}$]} & {\tiny $17.4 [\textcolor{red}{0.46}$]} \\
1 & {\tiny $0.0 [\textcolor{red}{0.00}$]} & {\tiny $0.1 [\textcolor{red}{0.00}$]} & {\tiny $0.0 [\textcolor{red}{0.00}$]} & {\tiny $0.0 [\textcolor{red}{0.00}$]} & {\tiny $0.1 [\textcolor{red}{0.00}$]} & {\tiny $0.0 [\textcolor{red}{0.00}$]} & {\tiny $0.0 [\textcolor{red}{0.00}$]} & {\tiny $0.0 [\textcolor{red}{0.00}$]} \\
2 & {\tiny $0.0 [\textcolor{red}{0.00}$]} & {\tiny $11.5 [\textcolor{red}{0.33}$]} & {\tiny $77.6 [\textcolor{red}{0.57}$]} & {\tiny $0.1 [\textcolor{red}{0.00}$]} & {\tiny $12.9 [\textcolor{red}{0.28}$]} & {\tiny $0.0 [\textcolor{red}{0.00}$]} & {\tiny $3.3 [\textcolor{red}{0.11}$]} & {\tiny $1.2 [\textcolor{red}{0.03}$]} \\
3 & {\tiny $0.0 [\textcolor{red}{0.00}$]} & {\tiny $0.0 [\textcolor{red}{0.00}$]} & {\tiny $2.4 [\textcolor{red}{0.04}$]} & {\tiny $93.2 [\textcolor{red}{0.19}$]} & {\tiny $70.6 [\textcolor{red}{0.52}$]} & {\tiny $0.1 [\textcolor{red}{0.00}$]} & {\tiny $0.0 [\textcolor{red}{0.00}$]} & {\tiny $0.1 [\textcolor{red}{0.00}$]} \\
4 & {\tiny $100.0 [\textcolor{red}{0.00}$]} & {\tiny $78.8 [\textcolor{red}{0.54}$]} & {\tiny $3.4 [\textcolor{red}{0.10}$]} & {\tiny $0.0 [\textcolor{red}{0.00}$]} & {\tiny $0.0 [\textcolor{red}{0.00}$]} & {\tiny $27.2 [\textcolor{red}{0.63}$]} & {\tiny $35.2 [\textcolor{red}{0.71}$]} & {\tiny $19.6 [\textcolor{red}{0.50}$]} \\
5 & {\tiny $0.0 [\textcolor{red}{0.00}$]} & {\tiny $0.0 [\textcolor{red}{0.00}$]} & {\tiny $0.0 [\textcolor{red}{0.00}$]} & {\tiny $5.4 [\textcolor{red}{0.15}$]} & {\tiny $4.7 [\textcolor{red}{0.11}$]} & {\tiny $36.6 [\textcolor{red}{0.78}$]} & {\tiny $9.4 [\textcolor{red}{0.25}$]} & {\tiny $2.2 [\textcolor{red}{0.06}$]} \\
6 & {\tiny $0.0 [\textcolor{red}{0.00}$]} & {\tiny $2.8 [\textcolor{red}{0.10}$]} & {\tiny $0.9 [\textcolor{red}{0.03}$]} & {\tiny $0.0 [\textcolor{red}{0.00}$]} & {\tiny $0.1 [\textcolor{red}{0.00}$]} & {\tiny $18.4 [\textcolor{red}{0.52}$]} & {\tiny $28.3 [\textcolor{red}{0.64}$]} & {\tiny $3.2 [\textcolor{red}{0.10}$]} \\
7 & {\tiny $0.0 [\textcolor{red}{0.00}$]} & {\tiny $0.0 [\textcolor{red}{0.00}$]} & {\tiny $7.1 [\textcolor{red}{0.20}$]} & {\tiny $1.2 [\textcolor{red}{0.03}$]} & {\tiny $10.1 [\textcolor{red}{0.22}$]} & {\tiny $3.3 [\textcolor{red}{0.10}$]} & {\tiny $11.6 [\textcolor{red}{0.35}$]} & {\tiny $55.0 [\textcolor{red}{0.78}$]} \\
8 & {\tiny $0.0 [\textcolor{red}{0.00}$]} & {\tiny $0.6 [\textcolor{red}{0.02}$]} & {\tiny $2.4 [\textcolor{red}{0.08}$]} & {\tiny $0.1 [\textcolor{red}{0.00}$]} & {\tiny $0.4 [\textcolor{red}{0.00}$]} & {\tiny $0.0 [\textcolor{red}{0.00}$]} & {\tiny $0.1 [\textcolor{red}{0.00}$]} & {\tiny $0.5 [\textcolor{red}{0.01}$]} \\
9 & {\tiny $0.0 [\textcolor{red}{0.00}$]} & {\tiny $6.2 [\textcolor{red}{0.18}$]} & {\tiny $5.6 [\textcolor{red}{0.17}$]} & {\tiny $0.0 [\textcolor{red}{0.00}$]} & {\tiny $1.2 [\textcolor{red}{0.02}$]} & {\tiny $14.3 [\textcolor{red}{0.41}$]} & {\tiny $5.9 [\textcolor{red}{0.15}$]} & {\tiny $0.8 [\textcolor{red}{0.01}$]} \\
\hline
\end{tabular}

\end{center}
\caption{Results for different rotations of an image.}
\label{tab:2}
\end{table}

\begin{table}[h]
\begin{center}
\begin{tabular}{l|llllllll}
\hline
{\small Labels} &
\begin{minipage}[b]{0.04\linewidth}{\vspace{.05cm}\includegraphics{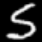}}\end{minipage} &
\begin{minipage}[b]{0.04\linewidth}{\includegraphics{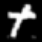}}\end{minipage} &
\begin{minipage}[b]{0.04\linewidth}{\includegraphics{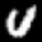}}\end{minipage} &
\begin{minipage}[b]{0.04\linewidth}{\includegraphics{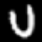}}\end{minipage} &
\begin{minipage}[b]{0.04\linewidth}{\includegraphics{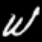}}\end{minipage} &
\begin{minipage}[b]{0.04\linewidth}{\includegraphics{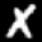}}\end{minipage} &
\begin{minipage}[b]{0.04\linewidth}{\includegraphics{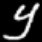}}\end{minipage} &
\begin{minipage}[b]{0.04\linewidth}{\includegraphics{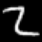}}\end{minipage} \\
\hline
0 & {\tiny $0.0 [\textcolor{red}{0.00}$]} & {\tiny $1.5 [\textcolor{red}{0.04}$]} & {\tiny $37.8 [\textcolor{red}{0.81}$]} & {\tiny $91.8 [\textcolor{red}{0.24}$]} & {\tiny $41.0 [\textcolor{red}{0.82}$]} & {\tiny $0.0 [\textcolor{red}{0.00}$]} & {\tiny $0.5 [\textcolor{red}{0.01}$]} & {\tiny $1.4 [\textcolor{red}{0.05}$]} \\
1 & {\tiny $0.0 [\textcolor{red}{0.00}$]} & {\tiny $3.1 [\textcolor{red}{0.09}$]} & {\tiny $0.0 [\textcolor{red}{0.00}$]} & {\tiny $0.0 [\textcolor{red}{0.00}$]} & {\tiny $0.0 [\textcolor{red}{0.00}$]} & {\tiny $21.7 [\textcolor{red}{0.54}$]} & {\tiny $0.0 [\textcolor{red}{0.00}$]} & {\tiny $0.2 [\textcolor{red}{0.00}$]} \\
2 & {\tiny $0.0 [\textcolor{red}{0.00}$]} & {\tiny $7.4 [\textcolor{red}{0.19}$]} & {\tiny $0.0 [\textcolor{red}{0.00}$]} & {\tiny $0.0 [\textcolor{red}{0.00}$]} & {\tiny $18.2 [\textcolor{red}{0.46}$]} & {\tiny $7.7 [\textcolor{red}{0.23}$]} & {\tiny $0.0 [\textcolor{red}{0.00}$]} & {\tiny $98.2 [\textcolor{red}{0.05}$]} \\
3 & {\tiny $2.1 [\textcolor{red}{0.07}$]} & {\tiny $1.7 [\textcolor{red}{0.05}$]} & {\tiny $0.0 [\textcolor{red}{0.00}$]} & {\tiny $0.0 [\textcolor{red}{0.00}$]} & {\tiny $0.1 [\textcolor{red}{0.00}$]} & {\tiny $0.0 [\textcolor{red}{0.00}$]} & {\tiny $38.4 [\textcolor{red}{0.80}$]} & {\tiny $0.2 [\textcolor{red}{0.00}$]} \\
4 & {\tiny $0.0 [\textcolor{red}{0.00}$]} & {\tiny $0.0 [\textcolor{red}{0.00}$]} & {\tiny $56.6 [\textcolor{red}{0.85}$]} & {\tiny $6.9 [\textcolor{red}{0.20}$]} & {\tiny $32.5 [\textcolor{red}{0.73}$]} & {\tiny $1.0 [\textcolor{red}{0.03}$]} & {\tiny $37.1 [\textcolor{red}{0.83}$]} & {\tiny $0.0 [\textcolor{red}{0.00}$]} \\
5 & {\tiny $97.9 [\textcolor{red}{0.07}$]} & {\tiny $0.8 [\textcolor{red}{0.03}$]} & {\tiny $0.0 [\textcolor{red}{0.00}$]} & {\tiny $0.0 [\textcolor{red}{0.00}$]} & {\tiny $1.0 [\textcolor{red}{0.04}$]} & {\tiny $0.8 [\textcolor{red}{0.02}$]} & {\tiny $19.7 [\textcolor{red}{0.54}$]} & {\tiny $0.0 [\textcolor{red}{0.00}$]} \\
6 & {\tiny $0.0 [\textcolor{red}{0.00}$]} & {\tiny $0.0 [\textcolor{red}{0.00}$]} & {\tiny $4.4 [\textcolor{red}{0.13}$]} & {\tiny $1.3 [\textcolor{red}{0.04}$]} & {\tiny $1.0 [\textcolor{red}{0.03}$]} & {\tiny $0.0 [\textcolor{red}{0.00}$]} & {\tiny $0.0 [\textcolor{red}{0.00}$]} & {\tiny $0.1 [\textcolor{red}{0.00}$]} \\
7 & {\tiny $0.0 [\textcolor{red}{0.00}$]} & {\tiny $13.8 [\textcolor{red}{0.34}$]} & {\tiny $0.0 [\textcolor{red}{0.00}$]} & {\tiny $0.0 [\textcolor{red}{0.00}$]} & {\tiny $0.0 [\textcolor{red}{0.00}$]} & {\tiny $0.0 [\textcolor{red}{0.00}$]} & {\tiny $3.6 [\textcolor{red}{0.10}$]} & {\tiny $0.0 [\textcolor{red}{0.00}$]} \\
8 & {\tiny $0.0 [\textcolor{red}{0.00}$]} & {\tiny $71.7 [\textcolor{red}{0.62}$]} & {\tiny $1.0 [\textcolor{red}{0.03}$]} & {\tiny $0.0 [\textcolor{red}{0.00}$]} & {\tiny $6.2 [\textcolor{red}{0.18}$]} & {\tiny $68.7 [\textcolor{red}{0.73}$]} & {\tiny $0.1 [\textcolor{red}{0.00}$]} & {\tiny $0.0 [\textcolor{red}{0.00}$]} \\
9 & {\tiny $0.0 [\textcolor{red}{0.00}$]} & {\tiny $0.1 [\textcolor{red}{0.00}$]} & {\tiny $0.1 [\textcolor{red}{0.00}$]} & {\tiny $0.0 [\textcolor{red}{0.00}$]} & {\tiny $0.1 [\textcolor{red}{0.00}$]} & {\tiny $0.0 [\textcolor{red}{0.00}$]} & {\tiny $0.6 [\textcolor{red}{0.01}$]} & {\tiny $0.0 [\textcolor{red}{0.00}$]} \\
\hline
\end{tabular}

\end{center}
\caption{Results for images from EMNIST dataset.}
\label{tab:3}
\end{table}

\section{Conclusion and future work}

We propose a simple and intuitive method, IGMC, that utilizes generative approaches to compute the posterior distribution of neural network predictions. We provide a convergence rate guarantee for IGMC and validate IGMC using a small convolutional architecture on the MNIST dataset.

IGMC still has many limitations. It needs a good $\Phi$ as a substitute for the prior and likelihood in the Bayesian method, but verifying whether $\Phi$ meets the requirements is not straightforward. IGMC requires generating $N\times H$ times, which can be computationally expensive, especially when evaluating the uncertainty of large neural networks since it involves training the generative model multiple times. We will study how to optimize IGMC in the future.


\medskip

\newpage



\newpage

\appendix

\section{Proof of Theorem~\ref{theorem:1}}
\label{appendix:proof}

Considering that $\hat F_{H,N}$ is the empirical distribution obtained by sampling $N$ observations followed by the distribution function $F_H$, our proof can be divided into the following two parts:

\subsection{Distance between $F_H$ and $F_\infty$}
\label{appendix:proof_1}

Let $\{y_k\}$ be a sequence of random variables satisfying $y_k\sim \Phi(\mathcal S,y_1,\ldots,y_{k-1})$, where $\mathcal S=x_1,\ldots,x_M$. Let $Z_K=\frac{\sum x+\sum y_k}{M+K}$, and $\mu=Z_H$.

Since $X$ in range $[0,1]$, we have $\mathrm{E}[|Z_K|]\leq 1$. And because there is $\mathrm E[\Phi(\mathcal S')]=\mathrm{mean}(\mathcal S')$ for any set $\mathcal S'$, obviously $\{Z_k\}$ is a martingale with respect to the filtration $\mathcal F_k=\sigma(y_1, \ldots, y_k)$, and the differences of $\{Z_k\}$ satisfies

$$|Z_k-Z_{k-1}|=\frac{1}{k+M}|Z_{k-1}+y_k|\leq\frac{2}{k+M}$$

According to the Azuma-Hoeffding inequality, following inequality holds for any $K>H$:

$$\mathrm{Pr}(|Z_K-\mu|\geq a)\leq 2\exp\left(-\frac{2a^2}{\sum_{k=H+1}^K \left(\frac{2}{k+M}\right)^2}\right)$$

When $K\to\infty$, we have

$$\lim_{K\to\infty}\mathrm{Pr}(|Z_K-\mu|\geq a)\leq2\exp\left(-\frac{H+M}{2}a^2\right)$$

We sample $\{y_k\}$ sequence for $S$ times, let $\{y^s_k\}$ be the sequence observed in the $s$-th sampling and $Z_K^s=\frac{\sum x+\sum y_k^s}{M+K}$. Then

$$\hat F^\Phi_{K,S}(t)=\frac{1}{S}\sum_{s=1}^S\mathbb I(Z^s_K<t)$$

is the empirical distribution by sampling $S$ times from $F_K$.

By the Glivenko–Cantelli theorem, when $S\to\infty$, $\sup_{t}|\hat F_{K,S}(t)-F_K(t)|\overset{a.s.}\to 0$, so

\allowdisplaybreaks
\begin{align*}
\left\|F_H(t)-F_\infty(t)\right\|_1
&=\int_0^1 \left|F_H(t)-F_\infty(t)\right|\mathrm{d}t\\
&=\int_0^1 \left|F_H(t)-\lim_{K\to\infty}F_K(t)\right|\mathrm{d}t\\
&=\int_0^1 \left|\lim_{S\to\infty}\left(\hat F_{H,S}(t)-\lim_{K\to\infty}\hat F_{K,S}(t)\right)\right|\mathrm{d}t\\
&=\int_0^1 \left|\lim_{S\to\infty}\frac{1}{S}\sum_{s=1}^S\left(\mathbb I(Z^s_H<t)-\lim_{K\to\infty}\mathbb I(Z^s_K<t)\right)\right|\mathrm{d}t\\
&\leq\lim_{S\to\infty}\frac{1}{S}\sum_{s=1}^S\int_0^1 \left|\mathbb I(Z^s_H<t)-\lim_{K\to\infty}\mathbb I(Z^s_K<t)\right|\mathrm{d}t\\
&=\lim_{S\to\infty}\frac{1}{S}\sum_{s=1}^S\left|\lim_{K\to\infty}Z^s_K-Z^s_H\right|\\
&=\mathrm{E}\left[\lim_{K\to\infty}\left|Z_K-Z_H\right|\right]\\
&=\int_0^1 \lim_{K\to\infty}\mathrm{Pr}(\left|Z_K-Z_H\right|>a)\mathrm{d}a\\
&\leq \int_0^1 2\exp(-\frac{M+H}{2}a^2)\mathrm{d}a\\
&<\sqrt\frac{2\pi}{M+H}
\end{align*}

\subsection{Distance between $\hat F_{H,N}$ and $F_H$}
\label{appendix:proof_2}

By Dvoretzky–Kiefer–Wolfowitz inequality, we have

\begin{equation}
\mathrm{Pr}\left(\sup_{x\in[0,1]}\left(\hat F^\Phi_{H,N}(t)-F^\Phi_H(t)\right)>a\right)\leq 2\exp(-2Na^2)
\end{equation}

Subsequently

\begin{align*}
\mathrm{E}\left[\left\|\hat F^\Phi_{H,N}(t)-F^\Phi_H(t)\right\|_1\right]
&=\int_0^1 a\cdot \mathrm{Pr}\left(\left\|\hat F^\Phi_{H,N}(t)-F^\Phi_H(t)\right\|_1=a\right)\mathrm{d}a\\
&=\int_0^1 \mathrm{Pr}\left(\left\|\hat F^\Phi_{H,N}(t)-F^\Phi_H(t)\right\|_1\geq a\right)\mathrm{d}a\\
&\leq\int_0^1 \mathrm{Pr}\left(\sup_{t\in[0,1]}\left(\hat F^\Phi_{H,N}(t)-F^\Phi_H(t)\right)\geq a\right)\mathrm{d}a\\
&\leq \int_0^1 2\exp(-2Na^2) \mathrm{d}a\\
&<\sqrt\frac{\pi}{2N}
\end{align*}

We complete the proof by combining the results of \ref{appendix:proof_1} and \ref{appendix:proof_2}. $\hfill\square$

\end{document}